\documentclass[journal,sort]{IEEEtran}

\usepackage{lineno,hyperref}
\usepackage{bm}
\usepackage{cancel}
\usepackage{booktabs}
\usepackage{multirow}
\usepackage{array,chngpage}
\usepackage{subfigure}
\usepackage{amssymb}
\usepackage{graphicx}
\usepackage{amsmath}
\usepackage{bm}
\usepackage{diagbox}
\usepackage{cite}
\usepackage{threeparttable}
\usepackage{url}
\usepackage{setspace}
\usepackage{caption}

\usepackage[linewidth=1pt]{mdframed}
\usepackage{lipsum}
\usepackage{booktabs}
\usepackage{color}
\usepackage{longtable}
\usepackage{doi}
\begin{document}

%

\title{Joint Learning of Deep Texture and High-Frequency Features for Computer-Generated Image Detection \thanks{This work is supported by Hong Kong Innovation and Technology Commission (InnoHK Project CIMDA), Hong Kong Research Grants Council (Project 11204821), City University of Hong Kong (Project 9610034).}
\thanks{Q. Xu,  H. Yan are with Department of Electrical Engineering, and
	Centre for Intelligent Multidimensional Data Analysis,
	 City University of Hong Kong, Kowloon, Hong Kong, China. (e-mail: qiangxu027@gmail.com,  h.yan@cityu.edu.hk).
 S. Jia is with  Department of Computer Science and Engineering, University at Buffalo (UB), State University of New York, NY, USA. (e-mail: shanjia@buffalo.edu). 
  X.-H. Jiang, T.-F. Sun are with School of Electronic Information and Electrical Engineering, Shanghai Jiao Tong University, Shanghai, China. (e-mail: \{xhjiang, tfsun\}@sjtu.edu.cn). 
 Z. Wang is with Department of Advanced Design
and Systems Engineering, and Centre for Intelligent Multidimensional Data Analysis,
City University of Hong Kong,
Hong Kong, China. (e-mail: Dr.Zhe.Wang@outlook.com). \emph{Corresponding authors: Zhe Wang and Shan Jia.} }}

\author{
Qiang~Xu,~\IEEEmembership{Member,~IEEE, }
Shan Jia,~\IEEEmembership{}
        Xinghao~Jiang,~\IEEEmembership{Senior Member,~IEEE,}
            Tanfeng~Sun,~\IEEEmembership{Senior Member,~IEEE,}
            Zhe Wang,~\IEEEmembership{Member,~IEEE}
	       Hong Yan,~\IEEEmembership{Fellow,~IEEE}
	    }

\maketitle

\begin{abstract}
%
Distinguishing between computer-generated (CG) and natural photographic (PG) images is of great importance to verify the authenticity and originality of digital images. However, the recent cutting-edge generation methods enable high qualities of synthesis in CG images, which makes this challenging task even trickier. To address this issue,  a joint learning strategy with deep texture and high-frequency features for CG image detection is proposed.
We first formulate and deeply analyze the different acquisition processes of CG and PG images. Based on the finding that multiple different modules in image acquisition will lead to different sensitivity inconsistencies to the convolutional neural network (CNN)-based rendering in images, we propose a deep texture rendering module for texture difference enhancement and discriminative texture representation.  Specifically, the semantic segmentation map is generated to guide the affine transformation operation, which is used to recover the texture in different regions of the input image. Then, 
the combination of the original image and the high-frequency components  of the original and rendered images are fed into a multi-branch neural network equipped with attention mechanisms, 
which refines intermediate features and facilitates trace exploration in spatial and channel dimensions respectively. 
Extensive experiments on two public datasets and a newly constructed dataset\footnote{The code and dataset are available at \url{https://github.com/191578010/TADA-NET.}} with more realistic and diverse images show that the proposed approach outperforms existing methods in the field by a clear margin. Besides, results also demonstrate the detection robustness and generalization ability of the proposed approach to postprocessing operations and generative adversarial network (GAN) generated images.
\end{abstract}


\section{Introduction}
\label{intro}

Nowadays, the advancement of digital imaging and media editing technologies have made it increasingly easier to
synthesize compelling computer-generated (CG) images. Although CG images have broadened the boundaries of digital multimedia, they have also aroused wide concerns about the authenticity of digital media \cite{swaminathan2008digital,xu2022motion} due to the fact that fake CG images may lead to misinformation and pose a serious threat to information ecosystem. One recent example is that Jonas Bendiksen, an award-winning documentary photographer fooled the world with computer-generated images, showing that the photojournalism industry is quite vulnerable to fake news pictures\footnote{Source: \url{https://www.magnumphotos.com/arts-culture/society-arts-culture/book-veles-jonas-bendiksen-hoodwinked-photography-industry/}.}.

Accordingly, recent years have seen a growing interest in the detection of CG image. 
Early studies 
focused on active forensics techniques, such as digital watermarking \cite{podilchuk2001digital, paquet2002wavelet}, where a watermark is embedded into the image before its delivery. 
The authenticity can be verified by comparing the extracted code with the original inserted code \cite{bouslimi2016crypto}. However, these techniques require a specific code to be embedded when the picture is taken and face the challenge of how to balance invisibility and robustness~\cite{panah2016properties}. 
Therefore, more and more researchers pay attention to passive (blind) forensics for fake image detection, which only relies on the intrinsic traces left in the digital media due to the image generation procedure \cite{xu2021detection, jiang2019detection}. 
Early attempts explored hand-crafted features or statistical analysis to represent the distinct intrinsic traces in CG and PG images, such as local binary pattern (LBP) \cite{li2012distinguishing}, color, texture, and shape features \cite{ke2013detecting}. 
However, the traditional features can hardly deal with complex images with heterogeneous origins \cite{quan2018distinguishing}. More recently, deep-learning approaches have shown strong performance in various vision applications as well as the CG image detection~\cite{tariang2019classification, dang2020tampered, yao2022cgnet,peng2022bdc}.
Though promising performance has been reported on specific
databases, deep-learning derived methods are well-known to
overfit when not trained with sufficiently representative data \cite{li2020face}. Furthermore, most deep features are automatically learned from the original images. How the distinct intrinsic traces are generated and how they can be effectively represented are ignored in the field of CG image detection due to the lack of deep exploration of the image acquisition process~\cite{guo2018fake, bappy2019hybrid, bayar2018constrained}.

In this paper, we try to have a clear understanding of how the CG and PG images are different in the generation process, and then explore more effective methods to enhance and represent the discriminative traces caused by the acquisition modules. Based on our findings that the texture differences between CG and PG images will be more significant after the convolutional neural network (CNN)-based rendering, we propose a novel data-driven approach for CG image detection. Our work makes the following contributions.

(1) A clear analysis of different acquisition processes in CG and PG images. The inherent traces of CG and PG images caused by different generation and processing operations are analyzed, which provide a basis for the trace learning strategy, and also make up for the lack of theoretical analysis in the current literature.

(2) A novel approach for deep texture representation. Different from the conventional straightforward strategy of hand-crafted or deep learning feature extraction, we design a texture rendering module to enhance the discriminative traces in images based on a semantic segmentation map-guided approach. 

(3) A joint learning strategy with deep texture and high-frequency features fusion. By adopting a separation-fusion detection strategy equipped with the attention mechanism, the proposed network can effectively learn the representative information of texture perturbation, high-frequency residual, and the global spatial trace in images. 

(4) Extensive evaluation with outstanding performance on a newly constructed dataset and two existing datasets. 
We collect a large-scale, highly-diverse, and realistic dataset with 6100 CG and 6100 PG images to address the limitations in existing datasets. Both intra-dataset and inter-dataset testing verify the superiority of the proposed method. Our method also shows outstanding robustness to postprocessing operations and generalization ability for GAN image detection.

The remainder of the paper is organized as follows. Section \ref{relawork} presents some recent related works on the detection of CG images. The differences between the CG image generation and the PG image acquisition processes are analyzed in Section \ref{anatheor}. Section \ref{proapp} describes the details of our proposed approach. Section \ref{exp} presents the experimental results and analyses.
Section \ref{conc} draws the main conclusions.
\section{Related Work}
\label{relawork}
The research on passive forensics of CG image detection can be categorized into two groups according to the feature extraction strategies, i.e., hand-crafted feature-based and deep learning-based. This section will review the relevant studies and analyze how our work extends the literature.
 
 \subsection{Detection Based on Hand-Crafted Features}
Several methods in this category propose to extract the abnormal statistical traces left by specific graphic generation modules and use threshold-based evaluation to detect computer-generated images. For example, 
Ng \emph{et al.} \cite{ng2005physics} revealed certain physical differences between the two image categories, such as the gamma correction in photographic images and the sharp structures in computer graphics, and then designed object geometry features for detection.
This method achieves a classification accuracy of 83.5\%, outperforming the cartoon features-based \cite{ianeva2003detecting} and the wavelet features-based method \cite{lyu2005realistic} (with accuracy of 71.0\% and 80.3\%, repectively). In \cite{wu2006detecting}, Wu \emph{et al.} utilized several visual features derived from color, edge, saturation, and texture features with the Gabor filter as discriminative features. Dehnie \emph{et al.} \cite{dehnie2006digital} emphasized that the image acquisition in a digital camera is fundamentally different from the generative algorithms deployed by computer-generated imagery. The properties of the residual image extracted by a wavelet-based denoising filter are designed for detection. After that, Pan \emph{et al.} pointed out that photo-realistic CG images are more surrealistic and smoother than natural images \cite{pan2009discriminating} and leveraged image perception characteristics to detect CG images. 

Another branch of methods in this category developed texture-based methods for automatic CG and PG classification. For example, Li \emph{et al.} used uniform gray-scale invariant local binary patterns in \cite{li2014distinguishing} and achieved 95.1\% accuracy with support vector machines.
Peng \emph{et al.} \cite{peng2015identification} combined 31-dimensional statistical and textural features to discriminate the acquisition pipelines of digital images. 
They further improved the performance by constructing a 9-D histogram feature and a 9-D multi-fractal spectrum feature to represent the distinct texture. 
In addition, the distribution of histogram \cite{wu2011identifying}, quaternion wavelet transform features \cite{wang2017forensics}, imaging and visual features \cite{zhang2011distinguishing}, etc., are used to detect CG images. 

Methods based on hand-crafted features provide credible theoretical interpretability, but they face the following two challenges: 1) the manual design of hand-crafted features can be tedious and limited by the capacity of feature description; 2) these features tend to have poor robustness to data with a high diversity in content, acquisition device, and forgery operation, etc.\cite{jia2020survey,gupta2021passive}.

\subsection{Detection Based on Deep Learning}
Deep learning-based methods have been overwhelmingly successful in computer vision, natural language processing, and video/image recognition \cite{chai2021deep, xu2018relocated} by automatical and adaptive feature learning. In the CG image detection task, a series of deep learning-based methods have also been proposed with promising performance. For example, Rahmouni \emph{et al.} \cite{rahmouni2017distinguishing} integrated a statistical feature extraction to a CNN framework to find the best feature for binary classification.
In \cite{yu2017identifying}, Yu \emph{et al.} constructed a convolution neural network trained on image patches and achieved an accuracy of 98.5\%. 
To further improve the performance, Quan  \emph{et al.} \cite{quan2018distinguishing} designed a network with two cascaded convolutional layers at the bottom of a CNN. The network can be easily adjusted to accommodate different sizes of input image patches while maintaining a fixed depth, a stable structure of CNN, and a good forensic performance. Yao \emph{et al.} combined the sensor pattern noise (SPN) with a patch-based five-layer model to detect CG images. 
Results show that the proposed model with three high-pass filters can achieve better results than that with only one or no filter \cite{yao2018distinguishing}. Zhang \emph{et al.} \cite{zhang2020distinguishing} proposed a CNN-based model with channel and pixel correlation. The key component of the proposed CNN architecture is a self-coding module that takes the color images as input to extract the correlation between color channels explicitly. 
Differently, the authors in \cite{meena2021distinguishing} designed a two-stream convolutional neural network. One stream uses a pre-trained VGG-19 network for trace learning, and the other stream preprocesses the images using three high-pass filters, which aim to help the network focus on noise-based distinct features of CG and PG images. 
A recent study in \cite{yao2022cgnet} proposed a network based on VGG-16 \cite{simonyan2014very} and Convolutional Block Attention Module\cite{woo2018cbam}, obtaining an accuracy of 96\% on DSTok dataset \cite{tokuda2013computer} after experimental validation.

Although promising CG image detection performance has been achieved by the aforementioned algorithms under the specific evaluation scenarios, three major shortcomings still exist for practical applications: 1) The inherent different generation mechanisms between CG and PG images are not described or fully considered in existing studies. 2) Features are directly learned from the original images, and no potential module is designed to drive the models to focus on discriminative traces. 3) Increasingly realistic CG images have made detection more difficult while there is a lack of large-scale and highly-diverse datasets for the evaluation of detection methods.

To remedy the shortcomings, we first formulate the differences in the acquisition between PG and CG images. Instead of directly learning features from the original images, we propose to use a rendering-based module to first enhance the discriminative features and further design a joint learning strategy with deep texture and high-frequency features for CG image detection. By employing a semantic segmentation map-guided deep texture rendering approach and adopting a separation-fusion detection strategy equipped with the attention mechanism, the proposed network can effectively learn the representative information of texture perturbation, high-frequency residual, and the global spatial trace in images. To fully evaluate the detection performance, we also present a new dataset named KGRA with a large
data size and high diversity in resolution range, lighting condition, image source, and image scene.
The dataset will be made publicly available to promote the development of CG image detection.

\section{Acquisition Processes of CG and PG Images}
\label{anatheor}
Due to the lack of a deep understanding of the inherent differences between CG and PG images, this section digs into their image acquisition processes and analyzes the discriminative traces that can be used for detection.
\begin{figure}[!ht]
\vspace{-0.3cm}
	\centering
	\includegraphics[width=0.42\textwidth,height=42mm]{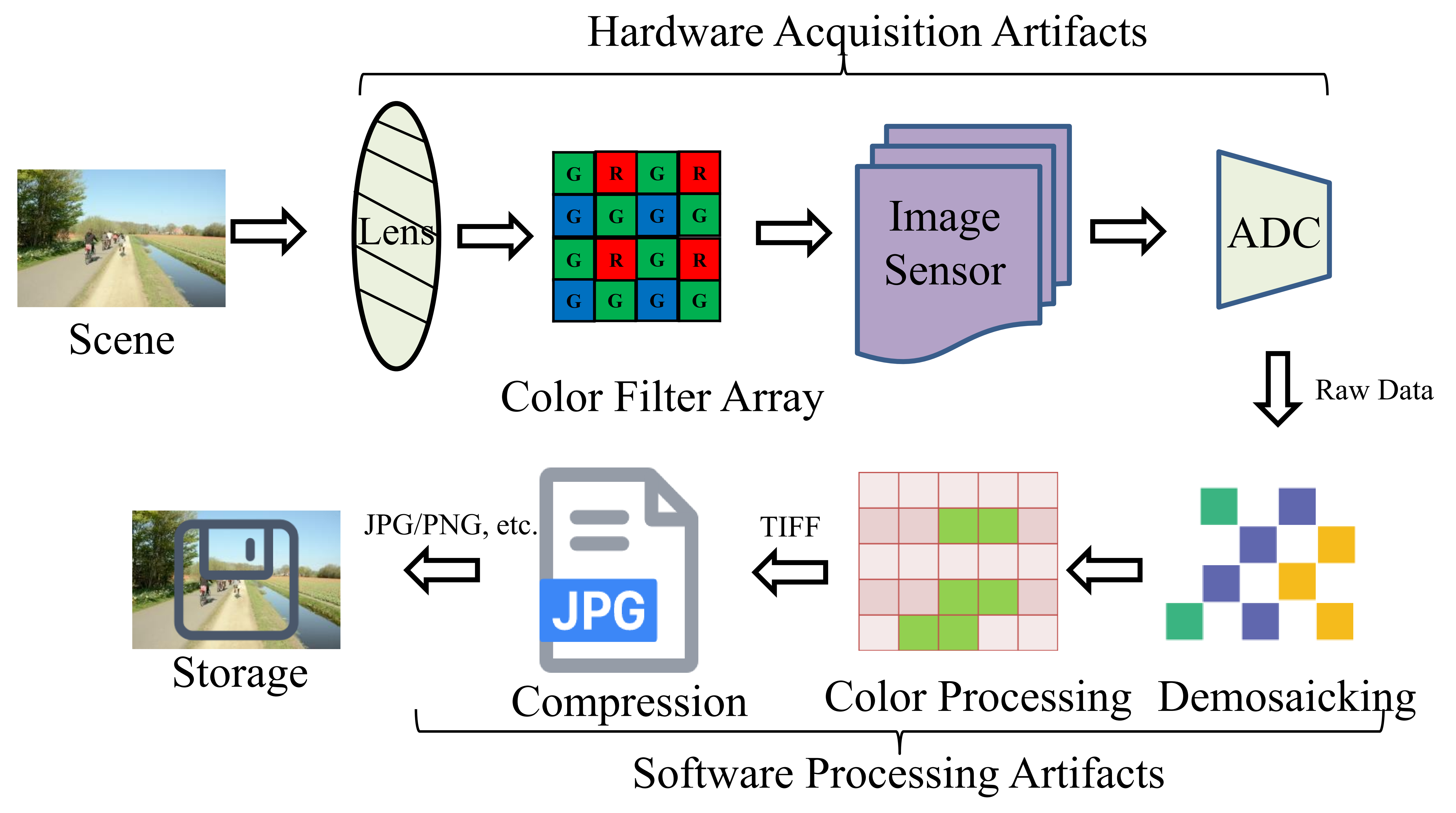}
	\caption{\label{fig:imgacq} Simplified diagram of PG image acquisition.}
\end{figure}


The acquisition of PG images has been widely analyzed~\cite{jinno2011multiple,lamoureux2005image}. As shown in Fig. \ref{fig:imgacq},  
the optical lens first conveys the light reflected from the scene towards the color filter array (CFA), which is a specific color mosaic that permits each pixel to gather only one particular light wavelength. Then, the output signal is sent to the imaging sensor (e.g., charged
coupled device, complementary metal oxide semiconductor), which is composed of an array of photo detectors, each corresponding to a pixel of the final image. After that, the analog-to-digital converter (ADC) converts the analog signal into digital form.
Further, demosaicing methods are applied to the raw data to conduct interpolation. For kernel-based interpolation method, the process can be expressed as $V_c(x,y)=\sum_{u,v=-N}^{N}h(u,v)\tilde{V}_c(x-u,y-v)$, where ${V}_c$,  $\tilde{V}_c$ represent the output and the original color signals, respectively. $c \in \{R, G, B\}$, $h(u, v)$ is the linear filter kernel function,  $N$ denotes the kernel size. After color values are recorded, more color processing operations (e.g., white balance, gamma correction) are conducted. Finally, the image data is compressed to reduce the cost of storage or transmission. 

\begin{figure}[!ht]
\vspace{-0.3cm}
	\centering
	\includegraphics[width=0.42\textwidth,height=45mm]{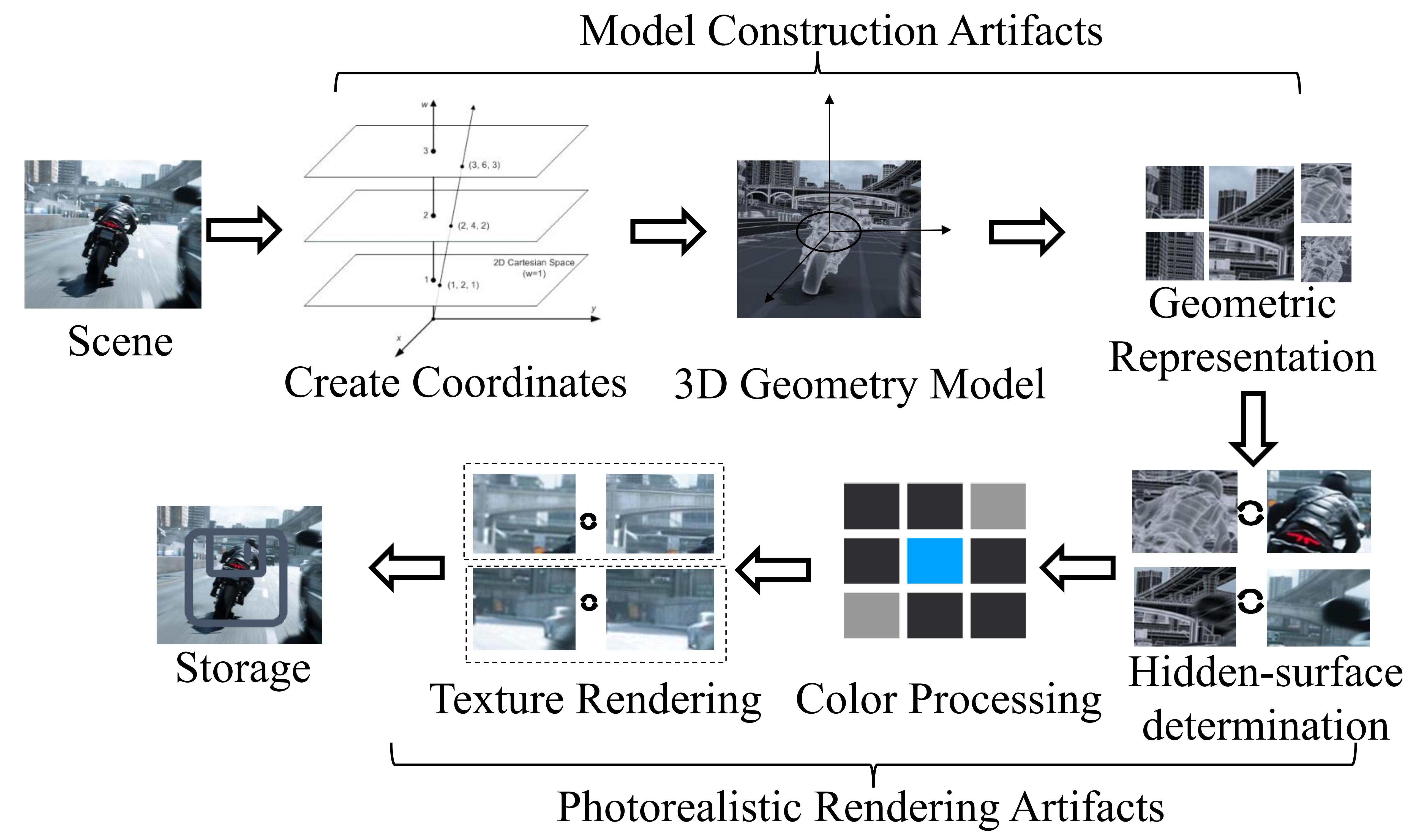}
	\caption{\label{fig:imgcgacq} Simplified diagram of CG image generation.}
\vspace{-0.3cm}
\end{figure}

We can observe that two levels of traces can be introduced in the generation of PG images, namely the hardware acquisition level (caused by lens, sensors) and software processing level (left by the CFA interpolation and other intrinsic image regularities). Therefore, we try to formulate the final traces in PG image as follows:
\begin{equation}
\mathbf{Tr}_{PG}=\mathbf{Tr}_{ha}+\mathbf{Tr}_{sp}
\end{equation}
where $\mathbf{Tr}_{ha}$ and $\mathbf{Tr}_{sp}$ denote the traces caused by hardware acquisition and software processing artifacts, respectively. In fact, traces in $\mathbf{Tr}_{ha}$ and $\mathbf{Tr}_{sp}$ have different representation forms. For example, the pattern traces in $\mathbf{Tr}_{ha}$ for a
specific camera with $I$ images can be estimated by: 
$\mathbf{Tr}_{pt}=1/I\sum_{i=1}^{I}\mathbf{PN}_i$, where 
$\mathbf{PN}_i=\mathbf{Img}_i-DN(\mathbf{Img}_i)$, and $\mathbf{PN}_i$ is the pattern noise of the $i^{th}$ image, $DN(\mathbf{\cdot})$ is the denoising operation. The lossy compression traces in $\mathbf{Tr}_{sp}$ can be denoted by $\mathbf{Tr}_{ct}=RT(IDT(\mathbf{D\times Q}))-IDT(\mathbf{D\times Q})$, where $RT(\cdot)$ is the rounding and truncation operations; $\mathbf{Q}$ and $\mathbf{D}$ are the quantization parameter and the coefficients, respectively; $IDT(\cdot)$ represents the inverse discrete cosine transform. All these traces will eventually affect the value of the image pixels in specific  areas.
\begin{figure}[!ht]
	\centering	\includegraphics[width=0.48\textwidth,height=42mm]{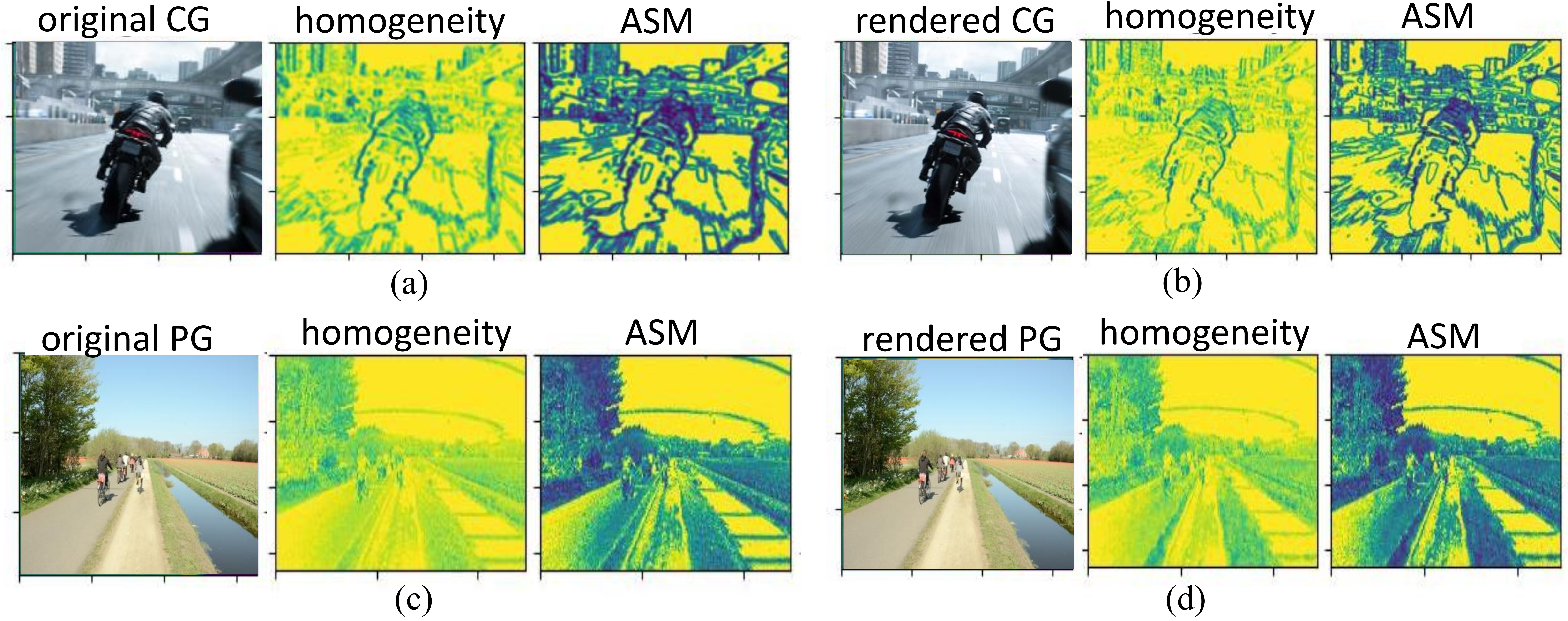}
	\caption{\label{fig:wenli} Comparison of GLCM-based feature maps for CG and PG example images before and after texture rendering. Figs. (a-d) are the results of the original CG, rendered CG, original PG and rendered PG images, respectively.} 
\vspace{-0.1cm}
\end{figure}

Different from the PG image acquisition process, CG images are directly generated by the computer-based algorithms for synthesis and manipulation. 
Based on the techniques in computer image generation~\cite{mustafa2019presenting}, we summarize the general pipeline of CG image generation in Fig. \ref{fig:imgcgacq}. Given one scene, the generation model  first establishes the coordinate system using computer aided design softwares~\cite{hall2012illumination}. In this step, the shape of the scene is expressed through a set of geometric data. Then, the geometric entities of an original design are modified to maximize performance and satisfy constraints via the representation of the geometric data. Next, a series of photorealistic rendering techniques are adopted to render the model, including eliminating the layers of objects that cannot be seen from the current viewpoint, establishing the illumination model, and correcting the color of different components~\cite{mustafa2019presenting}. Finally, the texture is rendered to make the picture more realistic, and the image data is further compressed for displaying. 

We further formulate the generation traces left by the generation procedure of CG images as
\begin{equation}
\mathbf{Tr}_{CG}=\mathbf{Tr}_{mc}+\mathbf{Tr}_{pr}
\end{equation}
where $\mathbf{Tr}_{mc}$ and $\mathbf{Tr}_{pr}$ denote the traces caused by model construction and photorealistic rendering artifacts, respectively. Similarly, the rendering trace in $\mathbf{Tr}_{pr}$ can be calculated as: $\mathbf{Tr}_{rt}=1/R\sum_{r=1}^{R}\mathbf{Ren}_r$, where $\mathbf{Ren}_r=\mathbf{Img}_r-RD{(\mathbf{Img}_r)}$, $R$ represents the total number of regions after semantic segmentation, and $\mathbf{Ren}_r$ is the rendering trace of the $r^{th}$ region of the image, $RD\mathbf{(\cdot)}$ is the rendering operation.

We can observe that the the acquisition process of CG images only contains the software-level processing, which generally aims to solve the following optimization problem:
\begin{equation}
\begin{aligned}
{CG}^{*}=\underset{CG}{argmin}\{d(\mathbf{ET}_{CG}, \mathbf{ET}_{PG})\}, \\
\emph{s.t.}\; d(\mathbf{Tr}_{CG},\mathbf{Tr}_{PG})<\varepsilon, \varepsilon>0 
\end{aligned}
\end{equation}
where ${CG}^{*}$ represents the output CG image,  $\mathbf{ET}_{CG}$, $\mathbf{ET}_{PG}$ denote the entities of CG and PG images, respectively. $\mathbf{Tr}_{CG}$ and $\mathbf{Tr}_{PG}$ denote the generation traces in the CG and PG images, respectively, and $d(\cdot)$ denotes the difference. 

From the above analysis, we can learn that due to the different generation modules, the inherent traces caused by both hardware acquisition and software processing exist in PG images, while the traces in CG images are mainly caused by software manipulation. The complicated software processing especially from modeling and rendering in CG images will inevitably suppress the high-frequency distribution of images and hardly replicate the attributes of
the high-frequency modes~\cite{dzanic2020fourier, chandrasegaran2021closer}. 
Therefore, the different acquisitions will lead to texture dissimilarity in the local repetitive patterns and their arranged rules in PG and CG images. 

To further show the dissimilarity, we 
present the Gray-Level Co-occurrence Matrix (GLCM)-based feature maps of CG and PG images with similar scenes in Fig. \ref{fig:wenli}. 
\emph{homogeneity} and \emph{Angular Second Moment (ASM)} calculated by the formulas $homogeneity=\sum_{i,j=0}^{N-1}\frac{P_{ij}}{1+(i-j)^2}$ and $ASM=\sum_{i=0}^{N-1}\sum_{j=0}^{N-1}{P_{ij}^2}$ are used to measure the smoothness and thickness of the gray level distribution of the image texture, respectively. $P_{ij}$ in the formulas represents the element of the normalized symmetrical GLCM. As can be seen from Figs. \ref{fig:wenli} (a) and (c), the local repetitive patterns and the arranged rules of the texture of CG image 
are rougher and more irregular than that of the PG image. 
To fully and automatically extract the discriminative texture patterns, we propose to use a CNN-based rendering (as introduced in Section \ref{proapp}) to perform the additional rendering for the two types of images to further enhance the difference. 
The feature maps of the rendered images are shown in Figs. \ref{fig:wenli} (b) and (d). It can be observed that after the rendering, the details of the content of these two types of images are enhanced to a certain extent. 
Due to the usage of geometric mapping methods and other computational techniques for processing, the rendering operation is proved to have different effects on the textures of the CG and PG images, which reveals a decrease in \emph{homogeneity} and \emph{ASM} of CG image, while the intensities of the PG image are enhanced to a certain extent. 
In summary, the sensitivity of CG and PG images to the CNN-based rendering is inconsistent, which motivates us to exploit this feature to explore inherent traces in CG images.
\begin{figure*}[!ht]
\vspace{-0.3cm}
	\centering
	\includegraphics[width=0.89\textwidth,height=65mm]{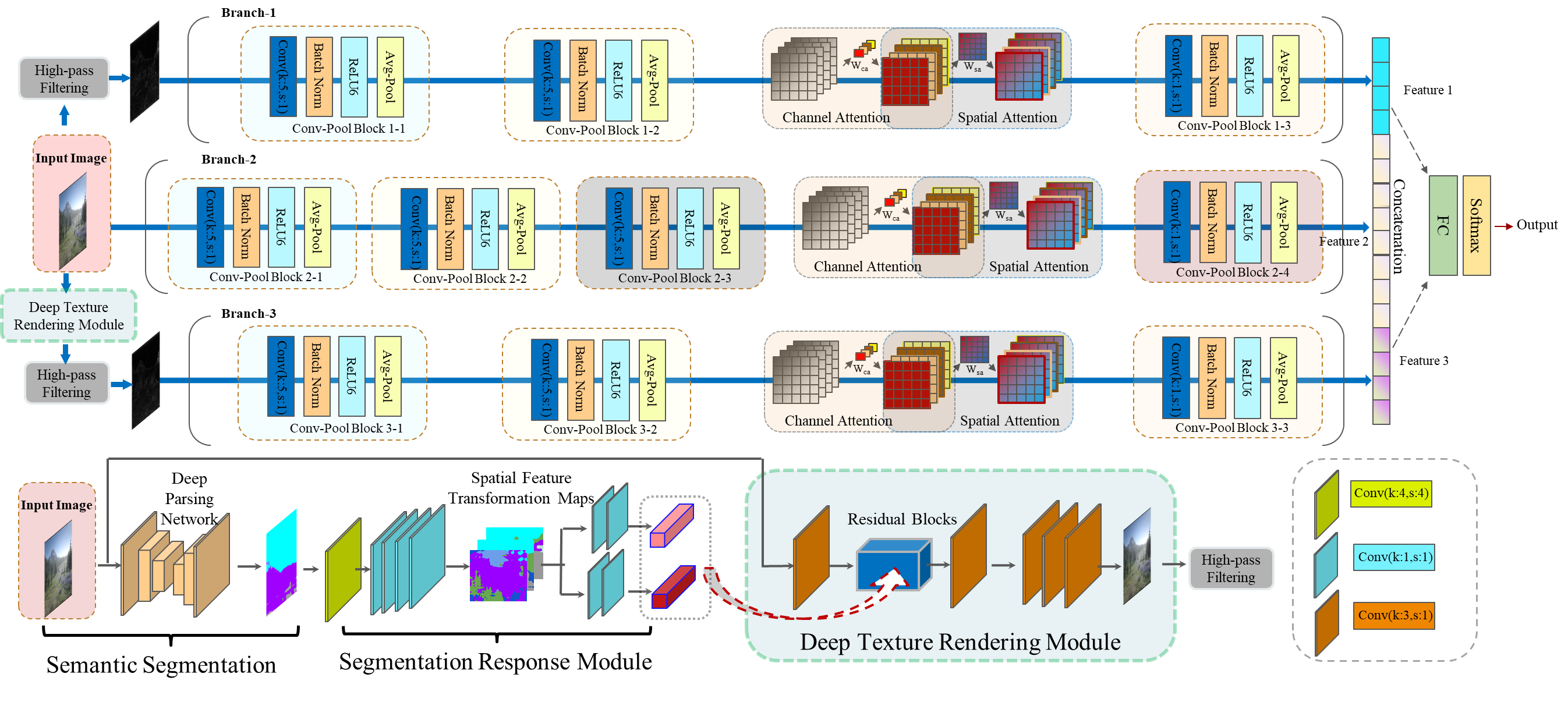}
	\caption{\label{fig:fram} The structure of the proposed detection network.  The input is an image of size 265 $\times$ 256, which is first fed into a semantic segmentation map-guided CNN-based deep texture rendering module for texture difference enhancement. The high-frequency components of the original image and the rendered image are obtained through high-pass filtering. Then, they are fed into a neural network with  three branches for automatic feature learning.  A softmax layer is added to the model to obtain the output probability of whether the input image is a PG or CG image.  }
	\vspace{-0.3cm}
\end{figure*}
\section{Proposed Approach}
\label{proapp}
Based on the analyses in Section \ref{anatheor}, the inherent differences in acquisition processes lead to different traces in CG and PG images. Besides, the inconsistency in the sensitivity of CG and PG images to CNN-based rendering results in different degrees of changes in their texture intensities, which show complementary properties.
These findings motivate us to adopt a joint learning strategy by combining features from different domains for the detection task. Specifically, we propose a novel data-driven approach by exploiting deep texture and high-frequency features learning to explore inherent traces left in CG images. In this section, the general structure of the proposed method is first presented. Then the component modules, including the deep texture rendering and the attention module, are introduced.

\subsection{General Structure}
\label{genstr}
To fully exploit the trace information to train a robust deep-learning model for CG image identification, we construct a hybrid deep-learning neural network that consists of three branches (branch-1, branch-2, branch-3) with heterogeneous structures. As shown in Fig. \ref{fig:fram}, we start by leveraging Deep Parsing Network \cite{liu2017deep} for semantic segmentation, the segmentation map is used to generate the spatial feature transformation and the produced feature maps, which are further forwarded to the deep texture rendering module for texture difference enhancement. After that, we extract the high-frequency components of the original and the rendered images to suppress the interference of the image contents. The two high-frequency components and the original image are fed into the three-branch network for classification. 
Note that we take the high-frequency components before and after rendering instead of the differences as the input of the network. This is because the difference information between these two components can be automatically learned by the network via branch-1 and branch-3, and such a strategy can better preserve other useful traces in the high-frequency domain.

Branch-1 and branch-3 of the network contain 3 Conv-Pool blocks, and branch-2 has 4 Conv-Pool blocks. These Conv-Pool blocks are with a convolution layer followed by an average-pooling layer. The two numbers in each Conv-Pool block name in Fig. \ref{fig:fram} represent the branch number and the index of the components, respectively. Channel and spatial attention schemes
are inserted into each branch to learn the representative information of the input image. Note that branch-1 and branch-3 are constructed for learning
the representative information of the original and rendered images in the high-frequency residual domain, respectively. Branch-2 is complementarily incorporated to learn the global representative information
of the input image. 
In short, the network can be denoted as follows:
\begin{equation}
\mathbf{NW}=\{ B_1(\mathbb{H}(\mathbf{Img})), B_2(\mathbf{Img}), B_3(\mathbb{H}(\Phi (\mathbf{Img})))\}
\label{netw}
\end{equation}
where $B_i$ ($i\in \{1, 2, 3\}$) represents the $i^{th}$ branch, $\mathbf{Img}$ is the input image, and $\mathbb{H}(\cdot)$, $\Phi (\cdot)$  denote the high-frequency  components extraction and image rendering operations, respectively.

Specifically, in our network, a convolution layer with 16 output
channels is performed on each processed image, followed by batch normalization (BN) \cite{ioffe2015batch} and rectified linear
unit 6 (ReLU6) \cite{zou2020ship} activation function. Batch normalization and ReLU6 are adopted to prevent the network from overfitting and to increase the nonlinearity of the network, respectively. We use ReLU6 instead of the ReLU in \cite{nair2010rectified} in each Conv-Pool block, because we try to limit the activation to a maximum size of 6. The element-wise function of ReLU6 can be formulated as $ ReLU6(i)=min(max(0,i),6)$, where $i$ is the input signal. 
Right after the three components, the average-pooling rather than the max-pooling layer is adopted to focus more on local characteristics \cite{boureau2010theoretical}, and facilitate the learning of computer-generated traces. The process can be expressed as:
\begin{equation}
\mathbf{FM}_{out}^{cp_{ij}}=AP(min(max(0,[Conv(\mathbf{FM}_{in}^{cp_{ij}})]),6))
\label{convpool}
\end{equation}
where $\mathbf{FM}_{out}^{cp_{ij}}$ and $\mathbf{FM}_{in}^{cp_{ij}}$ are the output and input feature maps of the $j^{th}$ Conv-Pool block in the $i^{th}$ branch. ${[\cdot]}$ represents extracting the elements after batch normalization, ${AP(\cdot)}$ is the average-pooling operation.

Following the average-pooling layer, one Conv-Pool block with one convolutional layer followed by an average-pooling layer is inserted in branch-1 and branch-3, while two blocks are added in branch-2. The numbers of filters for the two blocks are 32 and 64. All the convolutional layers in these blocks are with kernel sizes of $5\times5$ and strides of $1\times1$.  Since the generated intrinsic traces are less salient than the image content information, the attention mechanism,  which mainly consists
of two parts, is adopted to facilitate the learning of such trace information. The first part is a channel attention scheme ($\mathbf{ATc(\cdot)}$) that weights the input feature of different channels ($\mathbf{FM}_{in}^c$). 
The second part is the spatial attention scheme ($\mathbf{ATs(\cdot)}$), and we use a lightweight and flexible convolutional block attention module similar to the spatial attention module proposed in \cite{woo2018cbam} to improve the representation of interests. The process can be formulated as follows:
\begin{equation}
\mathbf{FM}_{out}^s=\mathbf{ATs}({\mathbf{ATc}({\mathbf{FM}_{in}^c}}))
\label{atte}
\end{equation}
where $\mathbf{FM}_{in}^c$ denotes the input of the channel attention submodule, and  $\mathbf{FM}_{out}^s$ represents the output of the spatial attention submodule.

After the attention module, another Conv-Pool block is added to each branch. The numbers of filters in the convolutional layers are 64, 128, and 64, respectively. We restrict the receptive field in this block by using $1\times1$ kernel sizes with strides of $1\times1$ for the convolutional layers. The detailed parameters of each layer are provided in the \emph{supplementary material}.

In terms of the output of the framework, as formulated in Eq. \ref{fcca}, the original image and the high-frequency components
are forwarded to the backbone network. Then the features of these branches
are concatenated for classification, and the classification results on whether the input image is a CG or PG can be reported in the form of 0 or 1 through a fully-connected layer and a softmax layer, where 1 represents CG image, and 0 denotes PG image.
\begin{equation}
\mathbf{Output}=SoM(FC(Concat(\mathbf{ft_1,ft_2,ft_3})))
\label{fcca}
\end{equation}
where $\mathbf{ft_1}$,$\mathbf{ft_2}$,$\mathbf{ft_3}$ are the features in $B_1$, $B_2$, $B_3$, respectively. ${FC(\cdot)}$ is the fully-connection operation. ${SoM(\cdot)}$ is the softmax operation, and its element-wise operation can be expressed as:
\begin{equation}
\sigma(z_i) = \frac{e^{z_{i}}}{\sum_{j=1}^2 e^{z_{j}}} \ \ \ for\ i=1,2
\label{softma}
\end{equation}
where $z_i$ denotes the $i^{th}$ element of the input sequence.

%

\subsection{High-frequency Component Extraction}
\label{prepro}
For the sake of suppressing the impact of image semantic content on the learning of generation trace information in CG and PG images. In this work, high-pass filtering is performed on the original and rendered images to obtain the high-frequency components. Due to the inconsistency of the image size in the real-world scenarios, we first rescale the input images to 256$\times$256, and then perform the filtering. It is worth noting that there are no additional lossy compression traces introduced in the scaling process.  This procedure can be
formulated as follows:
\begin{equation}
\mathbf{PImg}\!=\!\!HP(RS(\mathbf{Img}))\!\!
=\!IFT(FT(RS\!(\mathbf{Img}))\ast MK)\!
\label{prep}
\end{equation}
where $\mathbf{Img}$ and $\mathbf{PImg}$ respectively denote the input and the processed images. $ RS(\cdot)$ denotes the rescaling operation, and $HP(\cdot)$ represents the high-pass filtering operation, where Fourier transform ($FT(\cdot)$) is applied to the image, and performs convolution between the transformed image and the filtering mask ($MK$). The resultant image can be obtained by conducting inverse fourier transform ($IFT(\cdot)$) of the convoluted image.

The extracted high-frequency components are forwarded to branch-1 and branch-3, and the rescaled image is  forwarded to branch-2 in order to provide complementary and global representations for detection.

\subsection{Deep Texture Rendering Module}
\label{TextureRendering}
\begin{figure}[!ht]
\vspace{-0.3cm}
	\centering
\includegraphics[width=0.48\textwidth,height=20mm]{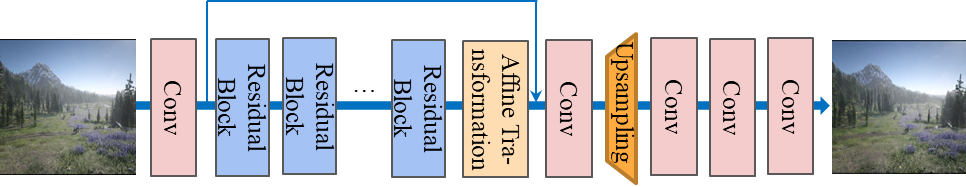}
	\caption{\label{fig:framsft}The structure of the deep texture rendering module. }
\end{figure}

\begin{figure}[!ht]
\vspace{-0.3cm}
	\centering
\includegraphics[width=0.42\textwidth,height=20mm]{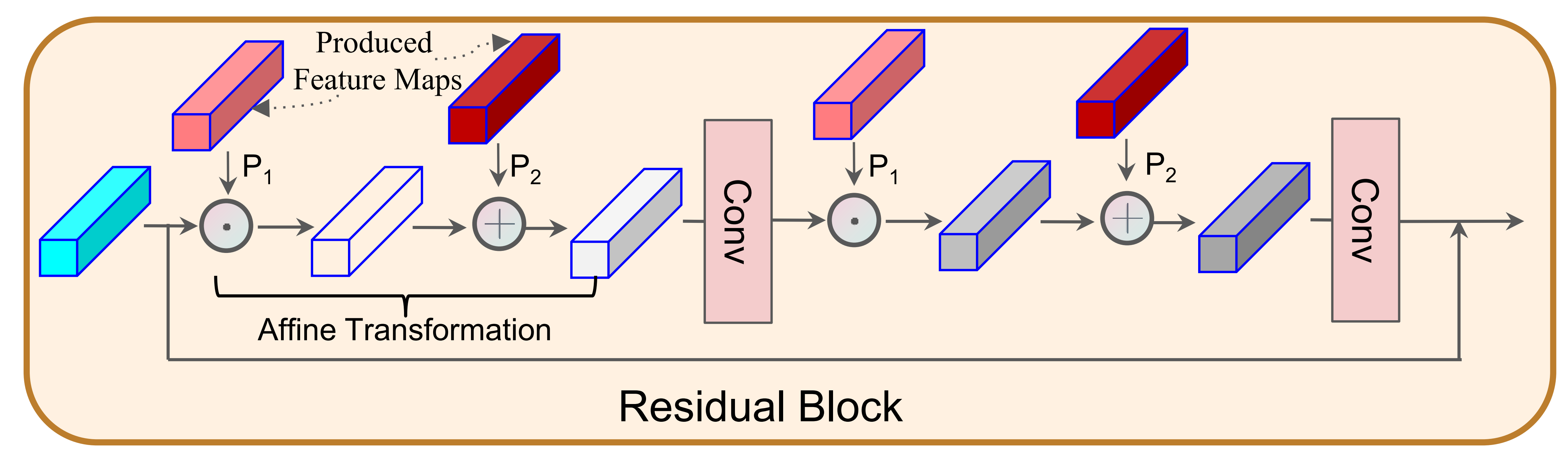}
	\caption{\label{fig:framresib}The illustration of the residual block. $P_1$ and $P_2$ represent the transformation parameters, which are obtained based on the produced feature map. }
\end{figure}

Since the sensitivity of CG and PG images to
the CNN-based rendering is inconsistent, it motivates us to 
develop a deep texture rendering module for texture difference enhancement. As shown in Fig. \ref{fig:framsft}, the module mainly contains convolutional layers
, semantic segmentation map-guided residual blocks, associated affine transformations and upsampling module. The input and output channels, kernel size, and stride of the first and last two convolutional layers are (3, 64, 3, 1) and (64, 3, 3, 1), respectively, while the remaining convolutional layers are set to (64, 64, 3, 1), the four numbers in parentheses represent the values of the corresponding parameters respectively.
The 16 residual blocks using a similar structure to the spatial feature transform network \cite{wang2018recovering} consist of two cascaded affine transformation submodules and two convolutional layers to perform feature-wise manipulation and spatial-wise transformation. The structure is shown in Fig.\ref{fig:framresib}, and its process can be formulated as follows:
\begin{equation}
\mathbf{FM}_{out}^r=Conv(AT(Conv(AT({\mathbf{FM}_{in}^r})))
\label{resfm}
\end{equation}
where $AT(\cdot)$ denotes the affine transformation, $\mathbf{FM}_{in}^r$, $\mathbf{FM}_{out}^r$ denote the input and the output of the residual block, respectively.

These blocks take the feature map from the previous layer as input, and applies two affine transformations, which modulate the feature map with a set of parameters ($P_1$, $P_2$) obtained by applying a learnable mapping function $MAP$ based on the segmentation maps $\Theta$, i.e., $MAP:\Theta\rightarrow(P_1, P_2)$. In short, the affine transformation can be formulated as follows:
\begin{equation}
AT(\mathbf{FM})=\mathbf{FM}\bigotimes P_1+P_2
\label{eqaffine}
\end{equation}
where $\bigotimes$ represents the element-wise multiplication, and $\mathbf{FM}$ is the feature map with the same dimension as $P_1$ and $P_2$.

\begin{figure}[!ht]
\vspace{-0.3cm}
	\centering
\includegraphics[width=0.42\textwidth,height=20mm]{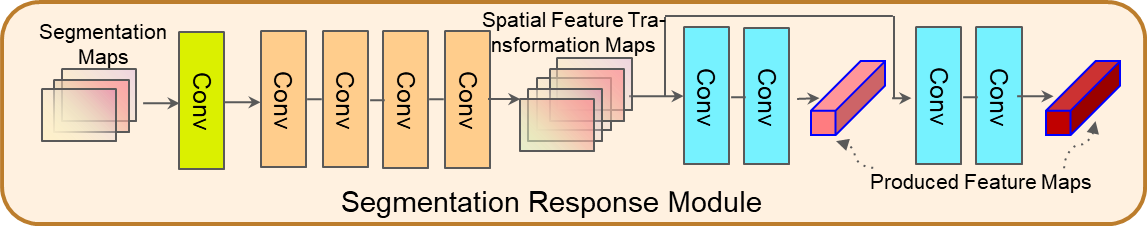}
	\caption{\label{fig:framrespon}The illustration of the segmentation response module.}
\end{figure}

We use semantic segmentation maps in the segmentation response module to obtain the produced feature map to guide the affine transformation operation. Specifically, we adopt the Deep Parsing Network \cite{liu2017deep} for semantic segmentation. The method incorporates high-order relations and a mixture of label contexts into the Markov Random Field (MRF), and it solves MRF by proposing a CNN that yields promising segmentation accuracies on several large-scale datasets.
 Then, we take the segmentation maps $\Theta$ as the input of the segmentation response module shown in Fig.\ref{fig:framrespon}, which are then fed to five consecutive convolutional layers. The generated intermediate spatial feature transformation maps are processed by two sets of convolutional modules, each containing two convolutional layers. Then, two produced feature maps are obtained, which are shared by the residual blocks. Note that the convolutional layers are with kernel sizes of 1$\times$1.

After the residual blocks and an additional affine transformation module, and following the convolutional layer, the nearest neighbor upsampling is employed at the back end of the rendering module. Besides, skip connection \cite{ledig2017photo} is adopted to ease the training process.

Similar to the strategy proposed in \cite{wang2018recovering}, adversarial learning is also adopted to make the rendered images as realistic as possible. A VGG-style \cite{simonyan2014very} network is used as the discriminator. The rendering module and the discriminator are jointly trained with an adversarial loss $L_{ad}=\sum_{i}log(1-D(G(x)))$ and a learning objective as follows:
\begin{equation}
\underset{G}{min} \underset{D}{max}V(D,G)=\underset{{y\sim p\textup{r}}}{\mathbb{E}}[log D(y)]+\underset{{x\sim p\textup{o}}}{\mathbb{E}}[log(1- D(G(x)))]
\end{equation}
where $p\textup{r}$ and $p\textup{o}$ represent the distributions of rendered and original images, respectively. $G(\cdot)$ denotes the rendering module.

In this way, texture in different regions of the image can be recovered. Since CG and PG images are generated in different ways, the rendering of textures will facilitate the enhancement of the texture differences, which in turn enables better detection of CG images.
\subsection{Attention Module}
 Attention mechanism has proven flexible and capable of capturing long-range feature interactions and boosting the representation capability of convolutional neural networks \cite{woo2018cbam}. In this work,  
 the channel attention submodule $\mathbf{ATc(\cdot)}$ and the spatial attention submodule $\mathbf{ATs(\cdot)}$ are connected in a sequential order, where the output of $\mathbf{ATc(\cdot)}$ will be the input of $\mathbf{ATs(\cdot)}$. More precisely, the input and output feature maps of the channel attention submodule are denoted as $\mathbf{FM}_{in}^c$, $\mathbf{FM}_{out}^c$, respectively. The channel attention submodule generates a 1-dimensional channel attention map $\mathbf{Mc(\cdot)}$, and perform element-wise multiplication on the input feature map. Therefore, $\mathbf{FM}_{out}^c$ can be formulated as $\mathbf{FM}_{out}^c=\mathbf{Mc}(\mathbf{FM}_{in}^c)\bigotimes \mathbf{FM}_{in}^c$, where $\bigotimes$ is the element-wise multiplication. Similarly, the output feature map of the spatial attention submodule $\mathbf{FM}_{out}^s$ can be calculated as $\mathbf{FM}_{out}^s=\mathbf{Ms}(\mathbf{FM}_{out}^c)\bigotimes\mathbf{FM}_{out}^c$, where $\mathbf{Ms(\cdot)}$ is a 2-dimensional spatial attention map.

\subsubsection{Channel Attention Submodule}
\begin{figure}[!ht]
	\centering
\vspace{-0.2cm}
	\includegraphics[width=0.45\textwidth,height=25mm]{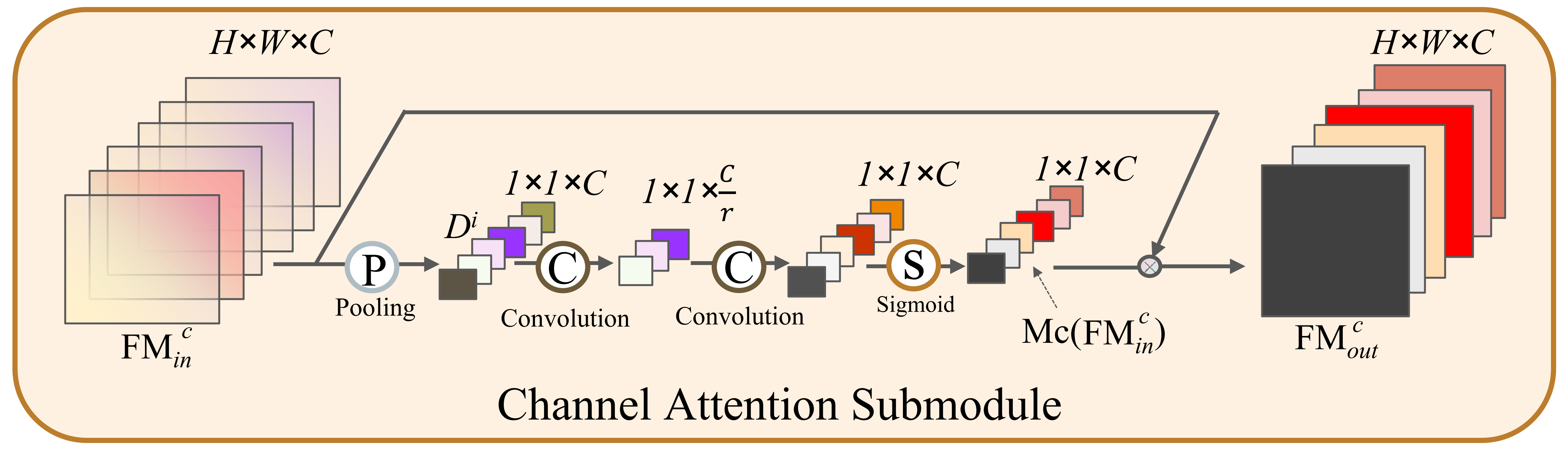}
	\caption{\label{fig:figure6a}The illustration of the channel attention submodule.}
\end{figure}

 Channel attention is added to the model to learn the weight of each channel. The illustration of the submodule is shown in Fig. \ref{fig:figure6a}. The input of the submodule is a combination of single feature maps with size $H \times W$, which can be represented by, $\mathbf{FM}_{in}^c=[{FM}_{in}^1, {FM}_{in}^2,..., {FM}_{in}^i,..., {FM}_{in}^C]$, $\mathbf{FM}_{in}^c$ in the formula denotes the input, $i$ and $C$ are the index and the total number of feature maps, respectively. These feature maps are squeezed into $C$ feature maps of size $1 \times 1$ in the spatial dimension by using global average-pooling. The $i^{th}$ element of the channel-wise statistic can be calculated as follows:
\begin{equation}
\mathbf{D}^{i}=\frac{1}{W\times H}\sum_{h=1}^{H}\sum_{w=1}^{W}{v}^{i}(h,w)
\label{eq2d1}
\end{equation}
where ${v}^{i}(h,w)$ is the value at position $(h, w)$ of ${FM}_{in}^i$.


Then, a convolution layer with a kernel size of $1\times1$ is used to perform channel-downscaling on channel-wise statistics. With the scaling ratio set to $r$, the output statistic of size $1\times1\times\frac{C}{r}$ can be obtained. Further, the downscaled statistic
is upscaled with ratio $r$ by putting through the second convolutional layer. After obtaining the recovered feature map of size $1\times1\times{C}$, it is fed into a sigmoid gate, which further outputs the final channel attention map. Finally, the input features are multiplied by the  final 1-dimensional channel weights $\mathbf{Mc(\mathbf{FM}_{in}^c)}$ to get the refined features $\mathbf{FM}_{out}^c$. In short, the channel attention procedure can be summarized as follows:
\begin{equation}
\mathbf{FM}_{out}^c=Sig(Conv(Conv(AP(\mathbf{FM}_{in}^c))))\bigotimes \mathbf{FM}_{in}^c
\label{eqfmins}
\end{equation}
where ${Sig(\cdot)}$ is the sigmoid function, which can be represented by $f_{sig}(x)=1/(1+e^x)$. ${Conv(\cdot)}$ denotes the convolution operation. ${AP(\cdot)}$ is the average-pooling operation.

\subsubsection{Spatial Attention Submodule}
\begin{figure}[!ht]
	\centering
	\includegraphics[width=0.42\textwidth,height=25mm]{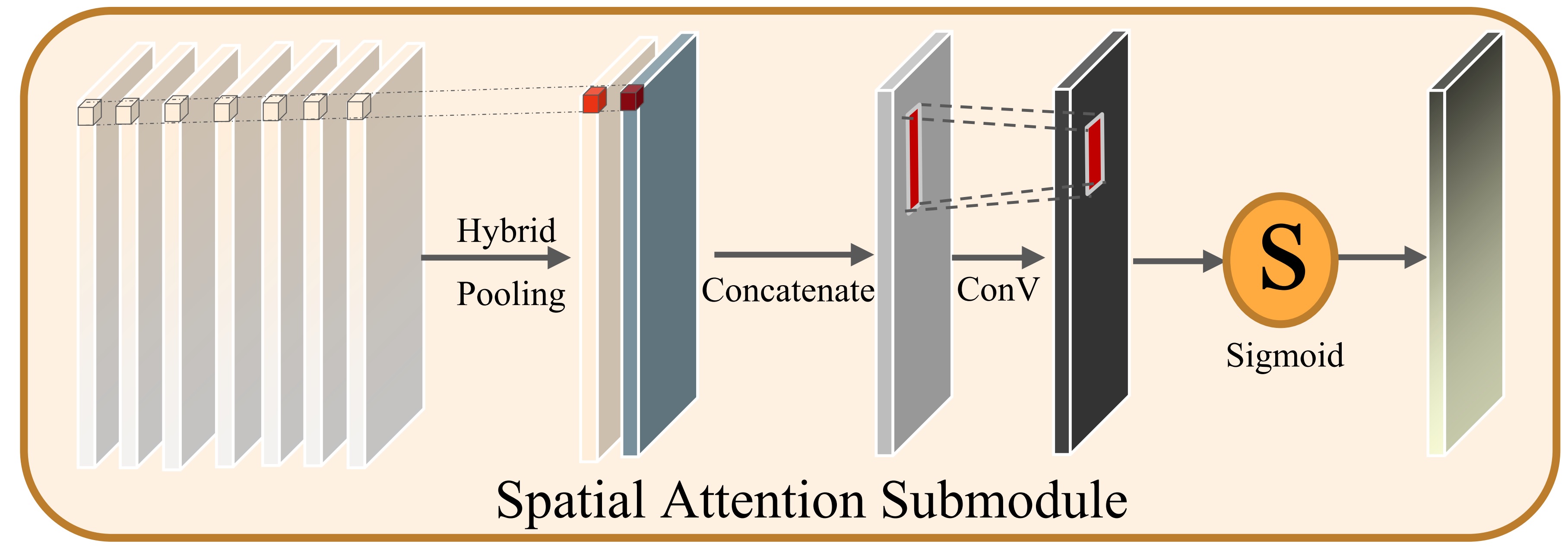}
	\caption{\label{fig:figure6b}The illustration of the spatial attention submodule.}
\end{figure}

Different from the channel attention submodule, the spatial attention submodule focuses on where is an informative part, which can provide information complementary to those of the channel attention mechanism. The illustration of the submodule is shown in Fig. \ref{fig:figure6b}. We first
employ a hybrid pooling operation consisting of an average-pooling and a max-pooling along the channel axis to summarize the average presence of the feature and the most activated presence of the feature, respectively. Then, we concatenate these two pooled features to generate an efficient feature descriptor, which is further convolved by a standard convolution layer. The output is also fed to a sigmoid gate in order to generate the 2-dimensional spatial attention map $\mathbf{Ms}(\mathbf{FM}_{out}^c)$. In real scenarios, the spatial attention map encodes where to emphasize or suppress. Similarly, the procedure can be summarized as:
\begin{equation}
\mathbf{Ms}(\mathbf{FM}_{out}^c)=Sig(Conv([AP(\mathbf{FM}_{out}^c), MP(\mathbf{FM}_{out}^c)]))
\label{eqfmouts}
\end{equation}
where ${AP(\cdot)}$ and ${MP(\cdot)}$ denote average-pooling and max-pooling operations, respectively.

Following the channel attention and spatial attention submodules, a Conv-Pool block is added to each channel. Finally, the classification results can be obtained through a fully-connected layer and a softmax layer.

\section{Experiments and Analyses}
\label{exp}


We conduct extensive experiments to evaluate the performance of the proposed approach in this section. First, the experimental
setup is introduced, including the datasets, evaluation criteria, and baseline approaches.
Then, we compare our algorithm with several state-of-the-art methods on different datasets. The parameter sensitivity is also measured to thoroughly evaluate the proposed approach. Moreover, we investigate the robustness against more realistic scenarios such as robustness on post-processed data. Since the images generated by Generative adversarial networks (GAN) are not naturally captured, we also verified the detection performance of the proposed approach for such images. All the experiments are conducted on a workstation with an Intel(R) Core(TM) i7-10700 CPU (2.90 GHz) processor, an NVIDIA GeForce RTX 3060 graphics card, and 64G DDR4 2666 MHz memory.
\begin{table}[!htbp]
\newcommand{\tabincell}[2]{\begin{tabular}{@{}#1@{}}#2\end{tabular}} 
\centering
\caption{Comparison of datasets for CG image detection.}
\resizebox{0.485\textwidth}{!}{
\begin{tabular}{|l|l|l|l|l|}
\hline
\multicolumn{2}{|l|}{Datasets}                              & KGRA (ours)                                                                                                                        & DSRah \cite{rahmouni2017distinguishing}                                                                                      & DSTok \cite{tokuda2013computer}\\
\hline
\multirow{2}{*}{Resolution}        & Min                    &500 $\times$ 281                                                                                                    & 1680 $\times$ 1050                                                                                         & 609 $\times$ 603        \\ \cline{2-5}
& Max  &4928 $\times$ 3264 &4928 $\times$ 3264 & 3507 $\times$ 2737 \\
\hline
\multirow{2}{*}{Number} & CG                     & 6100                                                                                                                         & 1800                                                                                       & 4850      \\\cline{2-5}
                                 & PG                     & 6100                                                                                                                         & 1800                                                                                       & 4850\\
\hline
\multirow{2}{*}{\tabincell{l}{Image\\Source}}    & CG                     & Kaggle,Google                                                                                                                & LDRD                                                                                       & Internet  \\\cline{2-5}
                                 & PG                     & \begin{tabular}[c]{@{}l@{}}RAISE (Nikon D40,D90,D7000 ),\\Personal Collection (Nikon D5200)\end{tabular} & \begin{tabular}[c]{@{}l@{}}RAISE(Nikon D40, \\D90, D7000~)\end{tabular} & Internet \\
\hline
\multicolumn{2}{|l|}{\multirow{2}{*}{Scene Coverage}}   & \multicolumn{3}{l|}{Outdoor, Indoor, Landscape, Nature, People, Objects, Buildings, Animal}                                                                                                                           \\\cline{3-5}
\multicolumn{2}{|l|}{}                                      & Light, Flame, Nighttime                                                                                                      & ~ ~ ~ ----                                                                 & --- \\
\hline
\end{tabular}}
\label{tab:datascategory}
\vspace{-0.5cm}
\end{table}

\subsection{Experimental Setup}
\label{Settingsdescription}
\textbf{Datasets.} The performance of the proposed approach is evaluated on a newly constructed dataset (named KGRA) with a high data diversity, and two public datasets (i.e., {DSRah dataset} \cite{rahmouni2017distinguishing} and {DSTok dataset} \cite{tokuda2013computer}).


\emph{KGRA dataset}: There are 6100 CG images and 6100 PG images sized from 500 $\times$ 281 to 4928 $\times$ 3264 with moderate
to good visual quality in the KGRA dataset. We construct our CG set by collecting CG video or game screenshots (namely Forza Horizon, GANTZ, God of War, Red Dead Redemption, and Playerunknown's Battlegrounds) in JPEG format from Kaggle and Google websites. 
For images in the PG set, 102 images with different contents were taken by the authors in Singapore using a NIKON D5200 camera, the focal length and exposure time are 25mm and {1}/{200} seconds, respectively. The rest images (5998 in total) were downloaded from the RAISE dataset, and directly converted from RAW format to JPEG. RAISE is primarily designed for the evaluation of digital forgery detection algorithms. All the images have been collected from four photographers, capturing different scenes and moments in over 80 places in Europe employing three different cameras \cite{dang2015raise}. Due to the inconsistent sizes of the images, we rescale them to 256 $\times$ 256 before feeding them into the network. We divided these images into a training set, a test set, and a validation set at a ratio of 4.1:1:1. 

\emph{DSRah dataset}\cite{rahmouni2017distinguishing}: This dataset consists of 1800 CG images downloaded from the Level-Design Reference Database \cite{Piaskiewicz} 
The author selected  five different video-game screenshots (i.e., Witcher 3, Battlefield 4, Battlefield Bad
Company 2, Grand Theft Auto 5, and Uncharted 4) to construct the dataset. The PG set is made up of 1800 natural images taken from the RAISE dataset  \cite{dang2015raise}.

\emph{DSTok dataset} \cite{tokuda2013computer}: This dataset contains 4850 CG images and 4850 PG images collected from the Internet. PG images include indoor and outdoor landscapes captured by different devices, and CG contents also contain different scenes. All the images are in JPEG format, and the file sizes are between 12 KB to 1.8 MB. 

The ratios of the samples used for training, validation, and testing on datasets DSRah and DSTok are 80\%, 10\%, and 10\%. The summary of the three datasets is shown in Table \ref{tab:datascategory}. It can be found that our KGRA dataset has a larger number of images and is more diverse than other datasets in terms of resolution range, image source and scene coverage.


\textbf{Baseline Approaches.} In this work, the proposed approach is compared with five popular CG detection algorithms with open-source codes. 

\emph{-- Histogram features-based method proposed by Wu {et al.}} \cite{wu2011identifying} (denoted as Wu-HIS), which employs the histogram bins of first-order and second-order difference images to distinguish CGs from PG images. 

\emph{-- Convolution neural networks constructed by Rahmouni {et al.} (namely Rah-CNN)} \cite{rahmouni2017distinguishing}, where a patch-based CNN with a custom pooling layer is designed. 

\emph{-- Convolutional neural networks-based method proposed by Quan {et al.} (namely Qu-CNN)} \cite{quan2018distinguishing}, which designs a model with a convFilter layer, three convolutional groups, two FC layers, and a softmax layer for different sizes of image patches.

\begin{table*}[htbp]
\caption{Comparison with existing methods on different datasets}
\vspace{-0.25cm}
\begin{center}
\setlength{\belowcaptionskip}{-0.2cm}
\resizebox{0.82\textwidth}{!}
	{
\begin{tabular}{|c|c|c|c|c|c|c|c|c|c|}
\hline
{Methods}&\multicolumn{3}{c|}{{KGRA Dataset}}&\multicolumn{3}{c|}{{DSRah Dataset}}&\multicolumn{3}{c|}{{DSTok Dataset}} \\\cline{2-10}

\textbf{} & {ACC (\%)}& TPR (\%)& {TNR} (\%) & {ACC} (\%)& {TPR} (\%)&{TNR} (\%)&ACC (\%)&{TPR} (\%)&{TNR} (\%)\\
\hline
Proposed& \textbf{97.42}&\textbf{96.98} & \textbf{97.86}& \textbf{97.03}&\textbf{96.78} & \textbf{97.28}& \textbf{96.58}&{96.15} & \textbf{97.01} \\
Wu-HIS \cite{wu2011identifying}& 85.62&85.18 & 86.06& 83.20&82.88 & 83.52& 82.42&82.30 & 82.54\\
Rah-CNN \cite{rahmouni2017distinguishing}& 88.50&89.18 &87.82 & 93.72&93.48 & 93.96& 75.88&75.05 & 76.71 \\
Zh-CPCR \cite{zhang2020distinguishing}& 92.83&92.41 &93.25 & 94.22&94.06 & 94.38& 92.28&92.04 & 92.52 \\
Qu-CNN \cite{quan2018distinguishing}& 94.33&93.81 &94.85 & 93.54&93.18 & 93.90& 93.02&92.88 & 93.16 \\
Yao-TLAM \cite{yao2022cgnet}& 94.76&94.28 &95.24 & 95.75&95.50 & 96.00& 96.10&\textbf{96.28} & 95.92 \\
\hline
\end{tabular}}
\label{tab1b}
\vspace{-0.2cm}
\end{center}
\end{table*}

\emph{-- Channel and pixel correlation-based method proposed by Zhang {et al.} (namely Zh-CPCR)} \cite{zhang2020distinguishing}, which analyzes channel and pixel correlation to model the differences between CG and PG images in terms of statistical characteristics.

\emph{-- Transfer learning with attention module-based method proposed by Yao {et al.} (namely Yao-TLAM)} \cite{yao2022cgnet}, where the authors introduce a CNN model to combine transfer learning and attention mechanism for CG image detection. 

\textbf{Other settings.} The proposed neural network is constructed using Pytorch-1.11.0 \cite{Pytorch} with CUDA-11.3 and Torchvision-0.12.0 package and trained with Adam optimizer \cite{kingma2014adam} of learning rate 0.0008. The cut-off frequency in high-frequency component extraction operation is assigned 30. The random horizontal and vertical flip probabilities are set to 0.3. We set the batch
size to 64, and the maximum number of epochs is fixed at 400. The cross-entropy loss is used to train the network.

\begin{figure}[!ht]
	\centering
\vspace{-0.4cm}
\setlength{\belowcaptionskip}{-0.2cm}
	\includegraphics[width=0.45\textwidth]{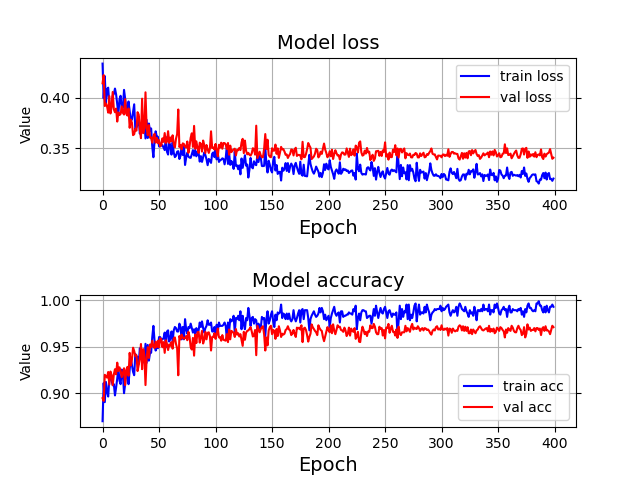}
	\caption{ The training and validation performance of the proposed network on the KGRA dataset.	}
	\label{fig: TRAVAL1}
\end{figure}
\textbf{Criteria.} To measure the detection performance, Accuracy (ACC), True Positive Rate (TPR), and  True Negative Rate (TNR) are used as the evaluation criteria, which can be expressed as: $ACC = \frac{TP + TN}{P + N}\times100\%$, $TPR = \frac{TP }{TP + FN}\times 100\%$ and $TNR = \frac{TN }{TN + FP}\times 100\%$,
where $TP$, $TN$, $FP$, and $FN$ in the formulas denote the number of true positive, true negative, incorrectly identified, and
incorrectly rejected  samples, respectively. $P$ and $N$ represent the number of positive and negative samples.


\subsection{Benchmark Experiments}
\label{Existing}
In this section, 
we compare the proposed method with five baselines on three CG image detection datasets, including the  KGRA, DSRah and DSTok datasets. 
we first show the variation of accuracy and loss with epoch during training and validation processes of the proposed method in Fig. \ref{fig: TRAVAL1}.
\begin{figure*}[!ht]
	\vspace{-0.2cm}
	\setlength{\abovecaptionskip}{-0.1cm}
	\setlength{\belowcaptionskip}{-0.1cm}
	\centering
	\includegraphics[width=0.93\textwidth]{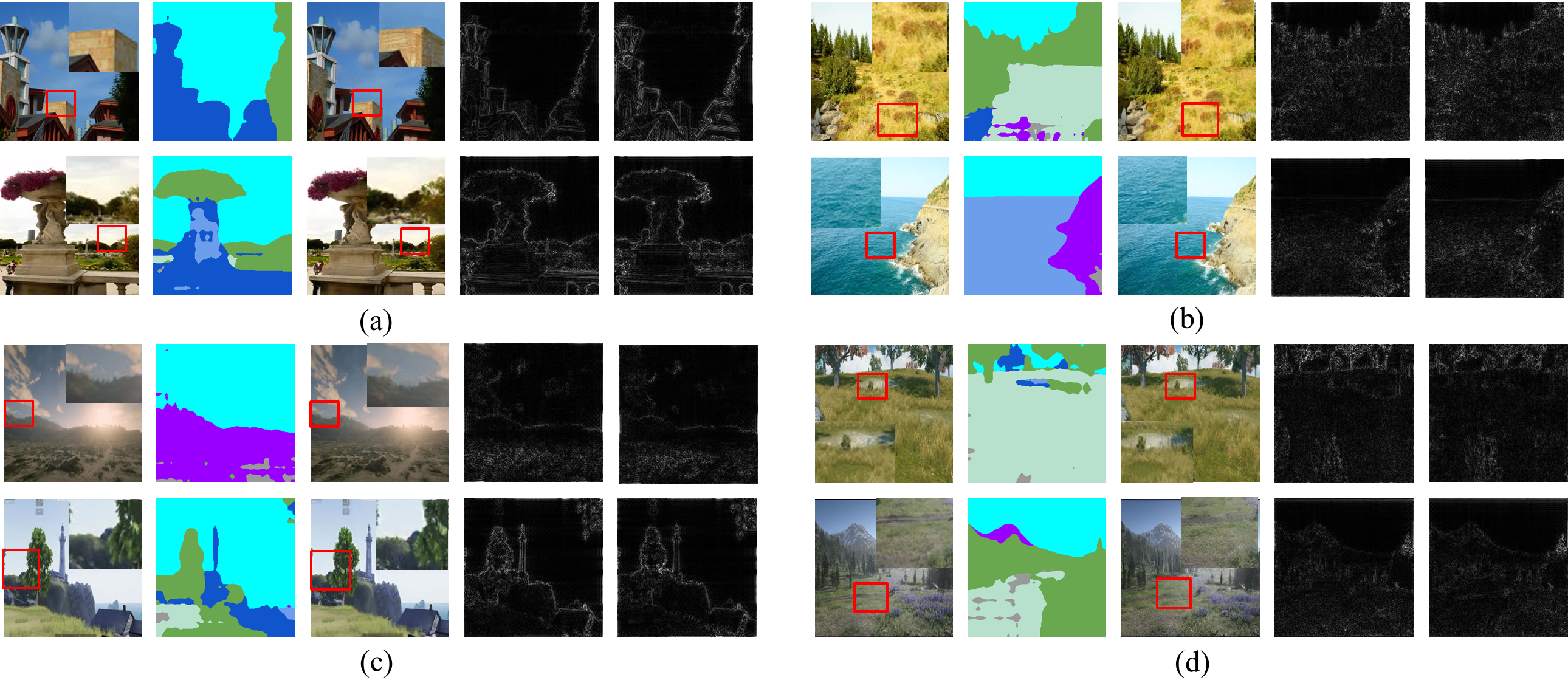}
	\caption{
		Visualization of deep texture rendering and the high-frequency components of the original and the rendered images. From left to right, each column in (a)-(d) represents the original images, the semantic segmentation images, the rendered images, and the high-frequency components of the original and the rendered images, respectively. The original images in (a) and (b) are from the PG dataset, while those in (c) and (d) are from the CG dataset. The corresponding zoomed version  of the contents marked with red boxes are also shown in the corner of the images.
	}
	\label{subfeature1t}
\end{figure*}

It can be observed from Fig. \ref{fig: TRAVAL1} that the training loss and validation loss decrease significantly in the first 150 epochs, and they plateau after around 200 epochs. The trend of accuracy rates is opposite to that of losses, which increases significantly in the first 150 epochs and stabilizes after 200 epochs. Compared with the verification process, the training process can ultimately achieve lower loss and higher accuracy. 

The comparison results are reported in Table \ref{tab1b}. It shows that the proposed approach achieves consistently higher accuracies on KGRA, DSRah, and DSTok datasets. For the performance on the KGRA dataset, the proposed algorithm  achieves the average accuracy rate
of 97.42\%, which is 11.80\%, 8.92\%, 4.59\%, 3.09\% and 2.66\% higher than those of \cite{wu2011identifying}, \cite{rahmouni2017distinguishing}, \cite{zhang2020distinguishing}, \cite{quan2018distinguishing} and \cite{yao2022cgnet}, respectively. Our approach also achieves better results on the KGRA dataset, achieving a TPR of 96.98\% and a TNR of 97.86\%. The average accuracies of our algorithm, \cite{wu2011identifying}, \cite{rahmouni2017distinguishing}, \cite{zhang2020distinguishing}, \cite{quan2018distinguishing} and \cite{yao2022cgnet} respectively reach 97.01\%, 83.75\%, 86.03\%, 93.11\%, 93.63\% and  95.54\%.
 The outstanding performance of our proposed method is reasonable since both texture enhancement and multi-branch feature fusion facilitate the model to effectively learn the discriminative trace information instead of focusing on image content information. 
 

\subsection{Sensitivity to Different Network Architecture Settings}

To further demonstrate the contribution of different modules in the proposed detection model, we show the impact of different parameters or network architecture settings on the detection performance on the KGRA dataset. 
Specifically, 
the following settings are considered to compare the detection performance: without high-pass filtering, no random horizontal/vertical flip, different arrangements of attention sub-modules (only channel, only spatial, spatial+channel), texture rendering disabled, and the use of Averaged Stochastic Gradient Descent (ASGD) optimizer. When changing one module, we keep other settings the same as those in Section \ref{Settingsdescription}. For the two cases of no high-pass filtering and no texture rendering, we build two branches for verification, where the original image is fed into one branch, and the high-frequency component and texture-rendered image are respectively forwarded to the other branch of the networks. 

\begin{table}[htbp]
\caption{Sensitivity to Different Network Architecture Settings on KGRA Dataset}
\begin{center}
\vspace{-0.2cm}
\setlength{\belowcaptionskip}{-0.1cm}
\resizebox{0.48\textwidth}{!}
	{
\begin{tabular}{|c|c|c|c|c|}
\hline
{Settings}&{Branch Number}&\multicolumn{3}{c|}{{Criteria(\%)}} \\
\cline{3-5}
\textbf{} & & {\textit{ACC}}& {\textit{TPR}}& {\textit{TNR}} \\
\hline
Proposed&Three& \textbf{97.42}&{96.98} & \textbf{97.86} \\
Without high-pass filtering &Two& 91.75&89.73 & 93.77\\
No random horizontal flip &Three& 96.80&96.64 & 96.96\\
No random vertical flip&Three& 96.93&\textbf{97.02} &96.84  \\
Spatial+Channel&Three& 97.14&96.87 &97.41  \\
Only Channel&Three& 94.35&94.10 &94.60  \\
Only Spatial&Three& 93.86&93.50 &94.22  \\
Texture rendering disabled&Two& 93.63&93.48 &93.78 \\
ASGD used as the optimizer&Three& 89.64&88.78 &90.50  \\
\hline
\end{tabular}}
\label{tab2dp}
\vspace{-0.2cm}
\end{center}
\end{table}

The comparison resulst are reported in Table \ref{tab2dp}. We can see that the detection accuracy without high-pass filtering achieves 91.75\%, which demonstrates the filtering operation can effectively capture high-frequency components, suppress the interference of the image contents, and facilitate the learning of discriminative traces.
For the scenario without random horizontal/vertical flipping, the detection accuracies reach 96.80\% and 96.93\%. These results imply that data augmentation can enhance the size and quality of training datasets such that better deep learning models can be built.  
Further, the performance decreases when the attention sub-modules are connected in other ways, and using two sub-modules at the same time outperforms using a single one. The better performance of the combination of modules proposed in our method can
be attributed to the stronger representation of inherent trace information. By leveraging channel and spatial attention sub-modules, the network is equipped with the ability to learn what and where to emphasize, which refines intermediate features and facilitates trace exploration effectively. We also notice that our approach using the Adam optimizer outperforms the method with the ASGD optimizer to some extent. This can be attributed to the stronger ability of the Adam optimizer in accelerating the convergence towards the relevant direction and reducing the fluctuation to the irrelevant direction in this detection task. 

Last but not least, the results also show a performance drop with depth texture rendering disabled. The finding infers that the submodule is helpful in improving the detection performance by providing refined and discriminative features. For the clarification of explanation, we demonstrate the effectiveness of depth texture rendering using the examples in  Fig.\ref{subfeature1t}, which shows the visualization of deep texture rendering and the high-frequency components of the original and the rendered images. As can be seen from the third column in each subfigure, after semantic segmentation, the objects of different categories in both CG and PG images are well segmented. Based on the segmentation map, the texture of the original image becomes more fine-grained after the rendering operation. After high-pass filtering, the high-frequency components of the rendered image and the original image show some differences, with the former shows more detailed information. In addition, it can be found that the intensities of the high-frequency components of the PG images in Figs.\ref{subfeature1t} (a-b) is increased significantly, while those in the CG images in Figs.\ref{subfeature1t} (c-d) show slight decrease. The finding reveals the evidence that the rendering module brings a considerable and complementary difference in high-frequency components of the CG and the PG images, which is crucial for the improvement of the detection ability of the proposed model.
\begin{table*}[htbp]
\caption{Performance evaluation with different postprocessing operations}
\vspace{-0.25cm}
\begin{center}
\setlength{\belowcaptionskip}{-0.2cm}
\resizebox{0.85\textwidth}{!}
	{
\begin{tabular}{|c|c|c|c|c|c|c|c|c|c|c|}
\hline
{Postprocessing}&{Methods}&\multicolumn{3}{c|}{{KGRA Dataset}}&\multicolumn{3}{c|}{{DSRah Dataset}}&\multicolumn{3}{c|}{{DSTok Dataset}} \\\cline{3-11}

&\textbf{} & {ACC (\%)}& TPR (\%)& {TNR} (\%) & {ACC} (\%)& {TPR} (\%)&{TNR} (\%)&ACC (\%)&{TPR} (\%)&{TNR} (\%)\\
\hline
&Proposed& \textbf{93.33}&\textbf{93.48} & \textbf{93.18}& {91.06}&{90.88} & 91.24& \textbf{92.08}&\textbf{92.25} & \textbf{91.91} \\
&Wu-HIS \cite{wu2011identifying}& 80.14&80.24 & 80.04& 78.58&79.42 & 77.74& 81.40&81.52 & 81.28\\
JPEG compression&Rah-CNN \cite{rahmouni2017distinguishing}& 84.65&84.33 &84.97 & 86.62&87.08 & 86.21& 73.62&72.85 & 74.39 \\
QF=95&Zh-CPCR \cite{zhang2020distinguishing}& 90.11&90.34 &89.88 & 91.37&91.05 & 91.69& 90.46&90.04 & 90.88 \\
&Qu-CNN \cite{quan2018distinguishing}& 89.63&91.42 &87.84 & 90.75&90.10 & 91.40&87.36&90.57 & 84.15 \\
&Yao-TLAM \cite{yao2022cgnet}& 91.87&91.63 &92.11 & \textbf{92.06}&\textbf{91.85} & \textbf{92.27}& 90.86&90.55 & 91.17 \\
\hline
&Proposed& \textbf{91.27}&\textbf{90.96} & \textbf{91.58}& \textbf{90.76}&\textbf{90.48} & \textbf{90.72}& \textbf{89.75}&{89.45} & \textbf{90.05} \\
&Wu-HIS \cite{wu2011identifying}& 78.72&78.68 & 78.76& 76.44&76.18 & 76.70& 78.66&78.84 & 78.48\\
JPEG compression&Rah-CNN \cite{rahmouni2017distinguishing}& 82.54&82.30 &82.78 & 85.66&85.10 & 86.22& 71.38&71.06 & 71.70 \\
QF=85&Zh-CPCR \cite{zhang2020distinguishing}& 87.38&87.59 &87.17 & 89.40&90.09 & 88.71& 87.18&87.34 & 87.02 \\
&Qu-CNN \cite{quan2018distinguishing}& 86.43&85.78 &87.08 & 88.70&87.98 & 89.42& 89.62&\textbf{90.10} & 89.14 \\
&Yao-TLAM \cite{yao2022cgnet}& 87.63&87.20&88.06 & 85.49&85.04 & 85.94& 86.17&85.79 & 87.61 \\
\hline
&Proposed& \textbf{87.58}&\textbf{87.24} & \textbf{87.92}& \textbf{86.43}&\textbf{86.78} & {86.08}& \textbf{85.50}&\textbf{85.27} & \textbf{85.72} \\
&Wu-HIS \cite{wu2011identifying}& 74.12&74.51 & 73.73& 73.48&72.96 & 74.00& 75.38&74.90 & 75.86\\
JPEG compression&Rah-CNN \cite{rahmouni2017distinguishing}& 79.69&79.48 &79.90 & 82.37&82.62 & 82.12& 68.37&67.85 & 68.89 \\
QF=75&Zh-CPCR \cite{zhang2020distinguishing}& 84.68&84.57 &84.79 & 86.33&85.86 & \textbf{86.80}& 83.85&83.54 & 84.16 \\
&Qu-CNN \cite{quan2018distinguishing}&82.46&84.47 &80.45 & 82.33&82.65 & 82.01& 83.16&83.28 & 83.04 \\
&Yao-TLAM \cite{yao2022cgnet}& 85.38&85.05 &85.71 & 84.46&84.28 & 84.69& 82.33&81.78 & 82.88 \\
\hline
&Proposed& \textbf{93.16}&\textbf{93.08} & \textbf{93.24}& \textbf{94.47}&\textbf{94.08} & \textbf{94.86}& \textbf{93.05}&\textbf{92.85} & {93.25} \\
&Wu-HIS \cite{wu2011identifying}& 67.38&65.48 & 69.28& 62.30&60.44 & 64.16& 71.63&71.89 & 71.37\\
Gaussian white noise&Rah-CNN \cite{rahmouni2017distinguishing}& 74.35&72.18 &76.52 & 78.58&78.41 & 78.75& 74.63&74.87 & 74.39 \\
Mean=0.01&Zh-CPCR \cite{zhang2020distinguishing}& 78.36&79.47 &77.24 & 79.45&80.37 & 78.53& 81.68&80.45 & 82.91\\
&Qu-CNN \cite{quan2018distinguishing}& 81.36&80.33 &82.39 & 77.68&74.82 & 80.54& 78.64&78.10 & 79.18 \\
&Yao-TLAM \cite{yao2022cgnet}& 91.30&91.08 &91.51 & 92.64&92.14 & 93.14& 92.88&92.37 & \textbf{93.39} \\
\hline
&Proposed& \textbf{92.02}&\textbf{92.48} & \textbf{91.56}& \textbf{93.17}&\textbf{93.50} & \textbf{92.84}& \textbf{92.69}&\textbf{92.30} & \textbf{93.08} \\
&Wu-HIS \cite{wu2011identifying}& 62.93&62.03 & 63.83& 58.46&55.37 & 61.55& 67.23&68.10 & 66.36\\
Gaussian white noise&Rah-CNN \cite{rahmouni2017distinguishing}& 71.49&72.44 &70.54 & 73.40&72.85 & 73.95& 68.53&66.57 & 70.49 \\
Mean=0.02&Zh-CPCR \cite{zhang2020distinguishing}& 73.33&72.50 &74.16 & 69.46&69.14 & 69.78& 75.68&74.24 & 77.12 \\
&Qu-CNN \cite{quan2018distinguishing}& 76.39&75.05 &77.73 & 73.15&69.88 & 76.42& 75.00&77.38 & 72.62 \\
&Yao-TLAM \cite{yao2022cgnet}& 90.26&89.78 &90.74 & 90.87&90.40 & 91.34& 91.43&90.98 & 91.88 \\
\hline
&Proposed& \textbf{94.12}&\textbf{94.28} & \textbf{93.99}& \textbf{94.63}&\textbf{94.88} & \textbf{94.38}& \textbf{93.72}&\textbf{93.55} & \textbf{93.89} \\
&Wu-HIS \cite{wu2011identifying}& 72.32&71.68 & 72.96& 70.58&72.11 & 69.05& 74.52&75.33 & 73.71\\
Salt and pepper noise&Rah-CNN \cite{rahmouni2017distinguishing}& 77.68&76.24 &79.12 & 85.45&85.98 & 84.92& 73.48&73.05 & 73.91 \\
Density=0.01&Zh-CPCR \cite{zhang2020distinguishing}& 82.35&80.40 &84.30& 80.33&81.17 & 79.49& 84.61&82.04 & 87.18 \\
&Qu-CNN \cite{quan2018distinguishing}& 85.03&84.33 &85.73 & 81.30&83.07 & 79.53& 83.47&86.08 & 80.86 \\
&Yao-TLAM \cite{yao2022cgnet}& 91.25&91.10 &91.40 & 92.34&92.60 & 92.08& 92.36&{92.08} & 92.64 \\
\hline
&Proposed& \textbf{92.85}&\textbf{92.48} & \textbf{93.24}& \textbf{91.53}&\textbf{91.72} & \textbf{91.34}& \textbf{93.08}&\textbf{92.84} & \textbf{93.32} \\
&Wu-HIS \cite{wu2011identifying}& 69.50&68.64 & 70.36& 72.39&71.48 & 73.30& 68.48&67.05 & 69.91\\
Salt and pepper noise&Rah-CNN \cite{rahmouni2017distinguishing}& 74.62&72.85 &76.39 & 81.36&80.66 & 82.06& 71.38&71.02 & 71.74 \\
Density=0.02&Zh-CPCR \cite{zhang2020distinguishing}& 77.68&75.76 &79.60 & 74.38&72.25 & 76.51& 78.89&76.30& 81.48 \\
&Qu-CNN \cite{quan2018distinguishing}& 79.15&79.28 &79.02 & 75.34&76.92& 73.76& 75.82&77.15 & 74.49 \\
&Yao-TLAM \cite{yao2022cgnet}& 88.56&88.23 &88.92 & 89.64&89.27 & 90.01& 91.28&{90.78} & 91.78 \\
\hline
\end{tabular}}
\label{postproc2}
\vspace{-0.1cm}
\end{center}
\end{table*}
 \subsection{Robustness to Postprocessing Operations}
 \label{datas}
 In real scenarios, forgers may conduct postprocessing operations, such as adding noise and image compression, to suppress the traces of CGs and deceive the forensics programs. Therefore, it is significant to investigate the robustness and sensitivity of the detection algorithm to different postprocessing operations. In this experiment, we conduct two types of postprocessing operations (i.e., JPEG compression and adding noise) on the CG images in the testing subsets. The quality factor (QF) is set to \{95, 85, 75\} to enable different quality levels. Two forms of noise often seen on digital images, namely Gaussian white noise and salt and pepper noise, are separately added to the images. We set the mean of Gaussian white noise and the density of salt and pepper noise to 0.01 and 0.02, respectively. More details about the visual comparison between the original image and the postprocessed images is provided in the \emph{supplementary material}. 

From the results in Table \ref{postproc2}, it can be observed that although different postprocessing operations are performed on the CG images, the detection accuracy of our approach is generally above 85\%. With the value of QF decreasing, detection accuracies of all approaches drop dramatically due to the heavy compression and loss by JPEG coding. 
Although the approach achieves the worst results when QF=75, our approach still outperforms other baselines 
in most cases, with average accuracies of 87.58\%, 86.43\% and 85.50\% on KGRA, DSRah and DSTok datasets, which are 1.20\%, 1.97\% and 3.17\% higher than the second-best approach \cite{yao2022cgnet}, respectively. It can also be observed that the proposed approach achieves higher accuracies when different noises are added, averaging 93.16\% on the KGRA dataset, 1.86\% higher than the state-of-the-art method \cite{yao2022cgnet} and 25.78\%  higher than the average accuracy of \cite{wu2011identifying}. With the increase of the mean of Gaussian white noise and the density of salt and pepper noise, the accuracies of all approaches  decrease. It can be attributed to the fact that a higher mean or density value reinforces the interference of noise information. This phenomenon agrees well with the early findings of Yao \emph{et al.} \cite{yao2022cgnet}. Interestingly, the results are slightly better than those when JPEG compression with QF fixed at 75. The main reason is that more information is lost during the compression process when QF is fixed at 75. It is also worth noting that the algorithm maintains outstandingly reliable detection accuracies of over 93.5\% in the scene where salt and pepper noise (Density=0.01) is added. The result proves that the proposed detection approach learns more robust and representative features. Overll, the proposed method demonstrates a strong robustness to different postprocessing operations.

\subsection{ Cross-dataset Evaluation}
 \label{crossdatas}
Considering that images from an unknown device may be used for testing in real scenarios, we evaluate the performance of our approach in the cross-dataset scenario. Since the testing set in the KGRA dataset does not overlap with the DSTok dataset, we therefore apply the model trained on the KGRA dataset to the testing set in the DSTok dataset. 
The comparison results are reported in Table \ref{tab1cross}.
\begin{table}[htbp]
\caption{Performance evaluation of the approaches in cross-dataset scenario}
\begin{center}
\setlength{\belowcaptionskip}{-0.2cm}
\resizebox{0.42\textwidth}{!}
	{
\begin{tabular}{|c|c|c|c|c|}
\hline
{Methods}&{Training Dataset}&\multicolumn{3}{c|}{{Testing Dataset: DSTok}} \\\cline{3-5}

\textbf{} &\textbf{} & {ACC (\%)}& TPR (\%)& {TNR} (\%) \\
\hline
Proposed&KGRA& \textbf{87.64}&\textbf{86.93} & \textbf{88.35}\\
Wu-HIS \cite{wu2011identifying}&KGRA& 65.05&63.78 & 66.32\\
Rah-CNN \cite{rahmouni2017distinguishing}&KGRA& 58.37&59.67 &57.07  \\
Zh-CPCR \cite{zhang2020distinguishing}&KGRA& 76.02&74.38 &77.66  \\
Qu-CNN \cite{quan2018distinguishing}&KGRA& 70.23&71.69 &68.77 \\
Yao-TLAM \cite{yao2022cgnet}&KGRA& 81.36&80.55 &82.17  \\
\hline
\end{tabular}}
\label{tab1cross}
\vspace{-0.4cm}
\end{center}
\end{table}

It can be observed from Table \ref{tab1cross} that our algorithm still achieves relatively promising performance in this scenario.  The detection accuracy of our approach reaches 87.64\%, outperforming the highest value achieved by the comparison algorithms with an accuracy of 81.36\%.
In terms of TPR and TNR, our algorithm achieves 86.93\% and 88.35\%, respectively, while the highest TPR and TNR of the comparison algorithm are 80.55\% and 82.17\%. This demonstrates the good generalization ability of the proposed method thanks to the rich and refined features learned by the designed texture enhancement module and joint feature learning strategy.

We also notice that the accuracies in the table  decrease significantly when compared with the results in Table \ref{tab1b}.  It is because the uniformity of training and testing data is not guaranteed, and the model can not satisfactorily distinguish images from the testing set merely using the information learned from the training set.  In summary, even though cross-dataset generalizability remains a challenging issue in CG image detection, 
the proposed algorithm outperforms the state-of-the-art algorithms in detecting unknown CG and PG images.

\subsection{Evaluation on GAN-Generated Image Detection}
 \label{gandatas}
A series of GAN-based methods have also been designed in the current literature to generate images from scratch as well as to modify the attributes of an existing image, which also posed a threat to the authenticity of digital. To show how our method can perform in detecting this kind of GAN-generated images, we conduct comparison results under both intra-dataset and cross-dataset testing scenarios. 
We first construct a GAN-generated image dataset (DSGGI), which contains 6100 images with different spatial resolutions generated by the pretrained StyleGAN2 model \cite{Karras2019stylegan2}. The contents of these images mainly contain \emph{car, cat, church, horse}. 
Then, we divided these images into a training set, a test set, and a validation set at a ratio of 4.1:1:1. 
The rest of the experiment settings and procedures are the same as those in Section \ref{Settingsdescription}. The models pretrained on the KGRA dataset and trained on the DSGGI dataset are used for testing, respectively. Results are reported in Table \ref{tab1crossga}.
\begin{table}[htbp]
\caption{Performance evaluation of the approaches for GAN-generated image detection.}
\begin{center}
\setlength{\belowcaptionskip}{-0.2cm}
\resizebox{0.48\textwidth}{!}
	{
\begin{tabular}{|c|c|c|c|c|c|c|}
\hline
{Methods}&\multicolumn{3}{c|}{{Training: KGRA/Testing: DSGGI}}&\multicolumn{3}{c|}{{Training: DSGGI/Testing: DSGGI}} \\\cline{2-7}

\textbf{}  & {ACC (\%)}& TPR (\%)& {TNR} (\%)& {ACC (\%)}& TPR (\%)& {TNR} (\%) \\
\hline
Proposed& \textbf{67.38}&{59.76} & \textbf{75.00}& \textbf{97.36}&\textbf{96.23} & \textbf{98.49}\\
Wu-HIS \cite{wu2011identifying}& 47.25&40.88 & 53.62& 75.68&70.74 & 80.62\\
Rah-CNN \cite{rahmouni2017distinguishing}& 42.68&39.47 &45.89& 78.47&74.20 &82.74  \\
Zh-CPCR \cite{zhang2020distinguishing}& 56.57&51.28 &61.91 & 86.52&85.50 &87.54 \\
Qu-CNN \cite{quan2018distinguishing}& 61.43&55.02 &67.84& 90.63&87.75 &93.51 \\
Yao-TLAM \cite{yao2022cgnet}& 62.40&\textbf{60.20} &64.60 & 94.50&92.85 &96.15 \\
\hline
\end{tabular}}
\label{tab1crossga}
\vspace{-0.4cm}
\end{center}
\end{table}

As shown in Table \ref{tab1crossga}, the proposed approach still achieves the best accuracies for GAN-generated image detection. Our method achieves
an average accuracy of 67.38\% under cross-domain testing, which is 20.13\%, 24.70\%, 10.81\%, 5.95\% and 4.98\% higher than those of \cite{wu2011identifying}, \cite{rahmouni2017distinguishing}, \cite{zhang2020distinguishing}, \cite{quan2018distinguishing} and \cite{yao2022cgnet}. Since the GAN images are generated in a different way, it is difficult for the model trained for CG image detection to accurately distinguish between GAN images and natural images. 
When the GAN-generated images are used for training,  our approach respectively outperforms \cite{wu2011identifying}, \cite{rahmouni2017distinguishing}, \cite{zhang2020distinguishing}, \cite{quan2018distinguishing} and \cite{yao2022cgnet} by 21.68\%, 18.89\%, 10.84\%, 6.73\% and 2.86\%, respectively. 

In addition, it can be observed that all algorithms achieved relatively low TPR values in this case. This indicates the networks are more inclined to misclassify GAN images as PG images. However, when the networks are trained on GAN-generated images, they can learn the difference information between the two types of images, resulting in a significant increase in accuracy. In summary, thanks to the deep texture rendering and the multi-branch feature learning mechanisms, the proposed network achieves outstanding performance even in GAN image detection. 

\section{CONCLUSION}
\label{conc}
In this paper, we have proposed a novel data-driven approach by exploiting texture enhanment and joint learning of deep texture and high-frequency features for CG image detection. By clearly formulating the different acquisition processes of PG and CG images, a deep texture rendering module is first designed for image texture representation and texture difference enhancement. Then we construct a multi-branch neural network equipped with the attention mechanism, in which the discriminative and rich traces of the original and rendered images are learned from the global spatial and high-frequency domains. 
A spatial and a channel attention module in each branch are developed to facilitate the network to learn more representative information. Evaluation on two public datasets as well as a more realistic and diverse dataset created by the authors demonstrate the promising performance of our approach.  In addition, the proposed approach is proved to be robust to different postprocessing operations, and also shows outstanding performance in GAN-generated image detection.
For the future work extending from this method, we will adopt more advanced learning algorithms to further improve the detection performance. We will also pay attention to CG image detection under more challenging light and color variants.

%
%

\bibliographystyle{IEEEtran}
\bibliography{mybibfile}

\end{document}